\documentclass[journal, onecolumn]{IEEEtran}

\usepackage[labelformat=simple]{subcaption}

\DeclareCaptionLabelFormat{subcaptionlabel}{\normalfont(\textbf{#2}\normalfont)}
\usepackage{algorithm}
\usepackage{algpseudocode}
\usepackage{amsfonts}
\usepackage{mathtools}
\usepackage{bm}
\usepackage{subcaption}
\usepackage{graphicx}
\usepackage{booktabs} 
\usepackage{soul}
\usepackage{xcolor}
\usepackage{hyperref}
\usepackage[numbers]{natbib}

\begin{document}

\title{Orientation-Constrained System for Lamp Detection in Buildings Based on Computer Vision}
\author{Francisco Troncoso-Pastoriza, Pablo Egu\'{i}a-Oller, Rebeca P. Díaz-Redondo, Enrique Granada-\'{A}lvarez, Aitor Erkoreka
\thanks{Francisco Troncoso-Pastoriza, Pablo Egu\'{i}a-Oller and Enrique Granada-\'{A}lvarez are with School of Industrial Engineering, University of Vigo, Campus Universitario, 36310 Vigo (Spain)}
\thanks{Rebeca P. Díaz-Redondo (rebeca@det.uvigo.es) is with School of Telecommunication Engineering, University of Vigo, Campus Universitario, 36310 Vigo (Spain)}
\thanks{Aitor Erkoreka (aitor.erkoreka@ehu.eus) is with ENEDI Research Group, Department of Thermal Engineering, University of the Basque Country, UPV, EHU, Alda. Urquijo , 48013 Bilbao, SpainSchool of Telecommunication Engineering, University of Vigo, Campus Universitario, 36310 Vigo (Spain)}
}

\maketitle

\abstract{Computer vision is used in this work to detect lighting elements in buildings with the goal of improving the accuracy of previous methods to provide a precise inventory of the location and state of lamps. Using the framework developed in our previous works, we introduce two new modifications to enhance the system: first, a constraint on the orientation of the detected poses in the optimization methods for both the initial and the refined estimates based on the geometric information of the building information modelling (BIM) model; second, an additional reprojection error filtering step to discard the erroneous poses introduced with the orientation restrictions, keeping the identification and localization errors low while greatly increasing the number of detections. These~enhancements are tested in five different case studies with more than 30,000 images, with results showing improvements in the number of detections, the percentage of correct model and state identifications, and the distance between detections and reference positions.}


\section{Introduction}
\label{intro}

Lighting is one of the most important aspects in the design, cost and maintenance of a building. Approximately, one-third of the electricity consumed in buildings corresponds to artificial lighting~\cite{Soori,Lombard,Baloch}, with a global demand that represents 19\% of all the electricity used in the world~\cite{iea2}. Recently, this consumption has increased at 2.4\% per year~\cite{iea2}. These~figures evidence the need for a more efficient use of the lighting resources, which requires a complete and precise inventory of the state of the lighting~elements.

The collected lighting information has to be integrated in the digital representation of the building, and building information modelling (BIM) is one of the most studied and used technologies to achieve this~\cite{Sanhudo,Rahmani}, integrating design and project data throughout the entire lifecycle of the building~\cite{Succar}. Formats~like industry foundation classes (IFC)~\cite{ifc} and green building XML schema (gbXML)~\cite{gbxml} provide the means to manage the digital representation of all the characteristics of the building, including lighting as one of the main aspects~\cite{Lu,Rahmani,Welle}. This is useful not only to store the new information of the lighting elements, but also to provide automatic systems with the necessary information to perform accurate detections of these elements in a given construction. In fact, the accuracy of the data is one of the main concerns in modelling and simulation~\cite{Lu}, and using the available BIM information as much as possible to improve this aspect should be a priority.

Automatic identification of lighting and lighting elements have been the focus of previous works. Elvidge~{et~al.}~\cite{Elvidge} analyzed the optimal spectral bands for the identification of lighting types and Liu~{et~al.}~\cite{Liu2} proposed an imaging sensor-based intelligent light emitting diode (LED) lighting system to obtain a more precise lighting control. Automatic detection systems can be implemented with different methods, but one of the most commonly used is computer vision. Computer vision systems (CVSs) can be applied to a wide variety of recognition problems; some of the methods use additional depth data, with RGB-D sensors or LiDAR technology, but the cost of the equipment required for these systems is higher and the performance can still be comparable or even better with methods that only rely on image information~\cite{D2CO}.

Lighting elements fall into the category of texture-less objects. The~detection and localization of this kind of objects is specially challenging, since the distinctive features that works on highly-textured objects are not present. Therefore, traditional object detection methods such as scale-invariant feature transform (SIFT)~\cite{SIFT} and speeded-up robust features (SURF)~\cite{SURF} do not work in this case. Thus, many alternatives have appeared to solve this problem based on edge information, and can be categorized into three groups: {keypoint-based}, {shape-based} and {template-based}.


First, keypoint-based algorithms, with a similar philosophy to SIFT and SURF, tries to generate descriptors incorporating invariant properties from keypoints in the image. Among these algorithms, Tombari~{et~al.} presented the bunch of lines descriptor (BOLD) features~\cite{Tombari}, using a compact and distinctive representation of groups of neighboring line segments. Later, Chan~{et~al.} developed the bounding oriented-rectangle descriptors for enclosed regions ({BORDER)~\cite{BORDER}, where they introduced a modified line-segment detection technique called {Linelets}, and more recently they introduced the binary integrated net descriptors (BIND)~\cite{BIND}, that encode multi-layered binary-represented nets for high precision edge-based description. Also in this category, Damen~{et~al.}~\cite{Damen1} used a system based on a constellation of edgelets, which was later improved~\cite{Damen2}.


The shape-based methods try to learn the shape of the object from the edge information. Ferrari~{et~al.}~\cite{Ferrari1, Ferrari2} introduced a method to learn the shape model of an object with a Hough-style voting for object localization. Moreover, Carmichael and Hebert~\cite{Carmichael} trained a classifier cascade to recognize complex-shaped objects in cluttered environments based on shape information.

Finally, the template-based methods were the first to provide good results for texture-less objects. Barrow~{et al.} first introduced the chamfer matching~\cite{Barrow}. Later, Borgefors~\cite{Borgefors} improved the correspondence measure and embedded the algorithm in a resolution pyramid, reducing the number of false matches and increasing the speed of the method. After that, many methods based on the chamfer matching appeared, including the work of Shotton~{et~al.}~\cite{Shotton}, introducing an automatic visual recognition system based on local contour features with an additional channel for the edge orientation. There is also the work of Hinterstoisser~{et~al.}~\cite{Hinterstoisser}, with a gradient-based template approach yielding faster and more robust results with respect to background clutter. Later, Liu~{et~al.}~\cite{fdcm} presented the fast directional chamfer matching (FDCM), using a joint location/orientation space to calculate the 3D distance transform, reducing its computational time from linear to sublinear. Based on this work, Imperoli and Pretto~\cite{D2CO} introduced the direct directional chamfer optimization (D$^2$CO) for object registration using the directional chamfer distance. The~works of Liu~{et~al.}~\cite{fdcm} and Imperoli and Pretto~\cite{D2CO} were used by Troncoso~{et~al.} to create a framework for the detection, identification and localization of lighting elements in buildings~\cite{Troncoso1}, which was later improved to work with any lamp shape~\cite{Troncoso2}.

These methods provide good detection results, but they do not fully utilize the available BIM information of the building, yielding intermediate values that are not tailored to the specific features of the given building space. Using the framework presented and later generalized in our previous works~\cite{Troncoso1,Troncoso2}, we tackle this problem by introducing two enhancements to the internal algorithms of the system: first, a new pose filter based on the reprojection error is used to discard erroneous poses; second, the optimization methods used in the initial estimation and in the refinement step are modified to force an orientation alignment based on the geometric data in the BIM of the building. These~modifications lead to better results in terms of the following three relevant metrics: total number of detections, identification performance, and distance to reference values.

The rest of the paper is structured as follows: Section~\ref{sec:method} contains the description of the methodology proposed in this work with the new additions to the system. Section~\ref{sec:exp} explains the experimental system used to evaluate the contributions, with the relevant results presented in Section~\ref{sec:results}. Finally, Section~\ref{sec:conclusions} contains the main conclusions of this work.

\section{Materials and Methods}
\label{sec:method}

The modifications proposed in this work were incorporated in the first part of the complete detection system~\cite{Troncoso1,Troncoso2}: the image and geometric processing, previous to the clustering and BIM insertion. The~updated diagram of steps is depicted in Figure~\ref{fig:general}. An in-depth description of this process can be found in~\cite{Troncoso1}, but we include here a brief explanation for the sake of completeness. The~first step corresponds to the extraction of initial pose candidates, where the image was analyzed to extract blobs and then obtain simplified shapes, either polygonal or elliptical. The~shapes were used to estimate candidate poses, and the set of candidates was filtered based on different thresholds. The~second step corresponds to the model selection, where the region of interest (ROI) was determined for each candidate, and the best model from the database was selected based on the fast directional chamfer matching (FDCM)~\cite{fdcm} computed in the ROI. Finally, in the last step the poses were refined based on the direct directional chamfer optimization (D$^2$CO)~\cite{D2CO}, and a score was obtained that is used to discard false positives.

\begin{figure}[h]
\centering
\includegraphics[width=.7\textwidth]{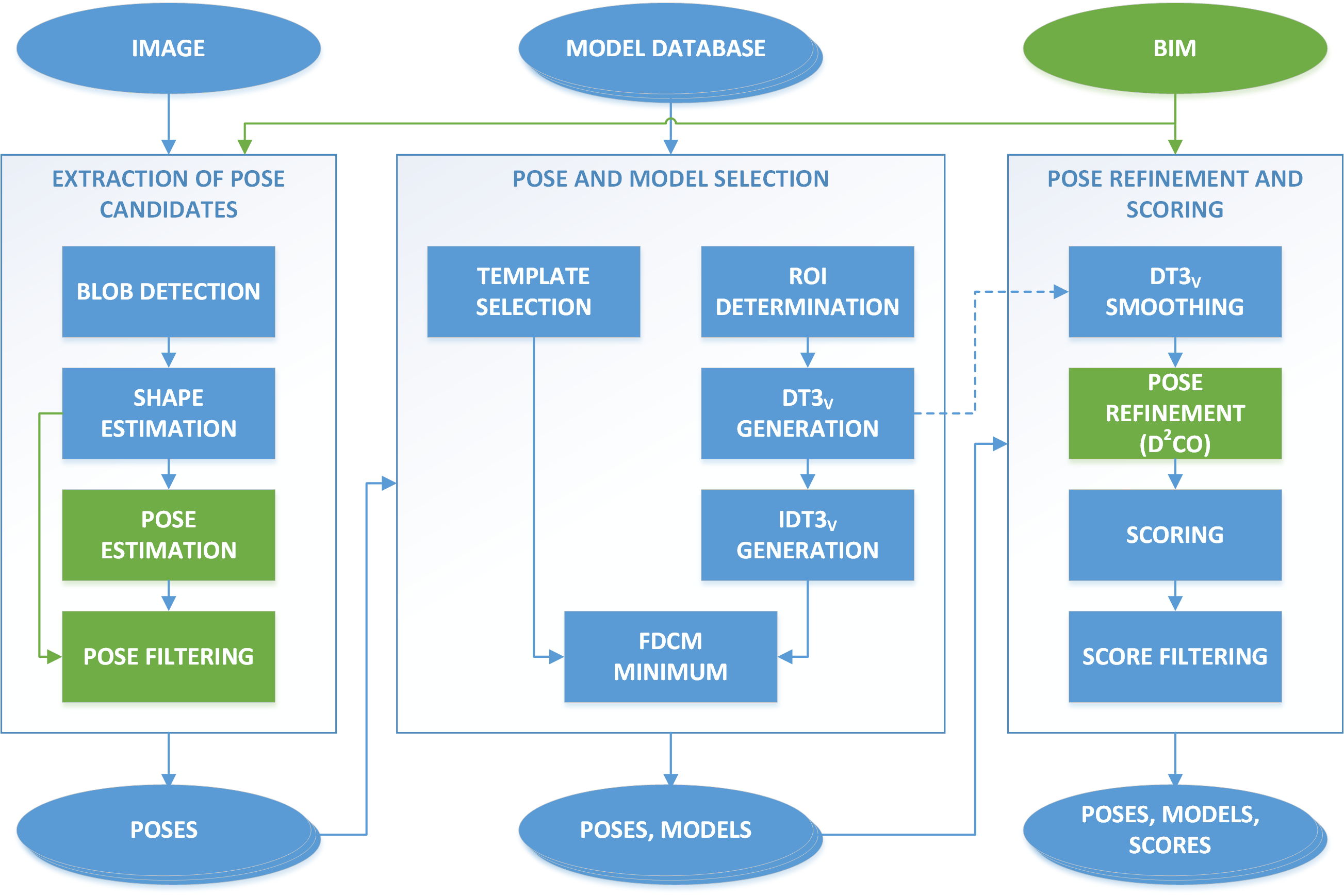}
\caption{Diagram of the modified detection system, with the inclusion of building information modelling (BIM) data in the first and third steps. The~modified parts with respect to~\cite{Troncoso1} are highlighted in green.}
\label{fig:general}
\end{figure}

The main contribution with respect to the system presented in~\cite{Troncoso1,Troncoso2} focuses on the use of BIM information of the building to align the orientations of the detections with the closest surface of the model. This introduces new restrictions in the extraction of pose candidates as well as the pose refinement, where both the initial pose estimation and the optimized version of this pose has to lie in a subset of the 3D rotation group.

This restrictions forced the estimates to be geometrically viable in much more cases than the original system. This was positive when the projection did not modify the original pose greatly, as it can compensate imprecisions in the original shape estimation, but it can also introduce errors when the modifications are too high, resulting in inaccurate detections that ended up falling inside the valid geometric limits. Thus, a new constraint was introduced to check the pose against the original light shape by means of the reprojection error.

The relevant altered steps to incorporate these enhancements are displayed in green in Figure~\ref{fig:general}. The~rest of this section describes in detail these modifications: the orientation alignment, that changes the pose estimation and pose refinement methods, and the pose filtering with the new reprojection error threshold.

\subsection{Orientation Alignment in Optimization Problems}
\label{sec:const}

The pose estimation and pose refinement steps were based on optimization problems that involve projections of points from model coordinates to homogeneous camera coordinates to evaluate their position in the image plane. Omitting the non-linear distortion effects for the sake of simplicity, this transformation process from a point $\bm p$ to a projected point $\bm p'$ is a well-known problem defined by the model $\bm M$, view $\bm V$ and projection $\bm P$ matrices, as presented in~(\ref{eq:mvp}):
\begin{align}\label{eq:mvp}
\bm{p}' = \bm{P} \bm{V} \bm{M} \bm{p} .
\end{align}


In the new system, we restricted the original problem to produce object poses that keep the orientation aligned with a given plane. This plane corresponds to the closest ceiling in the BIM model of the building in the case of embedded lamps, or the $z=0$ plane in the case of hanging lamps, as the ceiling orientation does not influence the orientation of the lamps for this type of models. This means that we needed to (a) force the initial orientation to have the $z$ axis parallel to the normal vector of the plane, and (b) restrict the possible orientation changes to only one degree of freedom, corresponding to rotations along this $z$ axis.

Let $\bm{w} \in \bf{\mathfrak{so}(3)}$ be the orientation vector of the {model} transformation $\bm{M}$, i.e., the transformation from model coordinates to world coordinates. Let $\bm{l} \in \bf{\mathfrak{so}(3)}$ be the orientation vector corresponding to a rotation that aligns the $z$ axis with the plane normal $\bm{\hat n}$, obtained as in Equations (\ref{eq:rvecp}) and (\ref{eq:rvec}), with a corresponding transformation matrix $\bm{L}$:
\begin{align}
\label{eq:rvecp} \bm{l}' = [0,0,1]^T \times \bm{\hat n}, \\
\label{eq:rvec}  \bm{l}  = \bm{l}' \frac{\arcsin \|\bm{l}'\|}{\|\bm{l}'\|} .
\end{align}

Then, to restrict the optimization problem we can transform the coordinate system inside the optimization problem and force the first two components of $\bm{w}$ to be zero, because a rotation vector of the form $[0,0,w_z]$ corresponds a rotation along the $z$ axis. The~required steps, depicted in Figure~\ref{fig:steps}, are as follows:

\begin{enumerate}
\item Calculate an aligned model matrix $\bm{M}^{(L)} = \bm{M} \bm{L}^{-1}$, with a corresponding orientation vector $\bm{w}^{(L)} = \left(w_x^{(L)}, w_y^{(L)}, w_z^{(L)}\right)$.
\item Project the vector $\bm{w}^{(L)}$ to the $z$ axis, setting the first two components to zero. The~result is $\bm{w}_p$, as presented in Equation (\ref{eq:wz}), with the corresponding transformation matrix $\bm{M}_p$, being $\bm{\hat z}$ a unit normal vector along the $z$ axis:
\begin{align}\label{eq:wz}
\bm{w}_p = \bm{\hat z} \left(\bm{\hat z} \bm{w}^{(L)}\right) = \left[0, 0, w_z^{(L)}\right]^T .
\end{align}

\item Use $\bm{w}_p$ in the optimization problem, constraining it to changes only in the third component of this vector and performing the projections as presented in (\ref{eq:lmvp}):
\vspace{12pt}
\begin{align}\label{eq:lmvp}
\bm{p}' = \bm{P} \bm{V} \bm{M}_p \bm{L} \bm{p} .
\end{align}

Thus, the degrees of freedom for the problem were reduced from six to four.

\item Calculate the final optimized model pose from the result of the optimization process, $\bm{M}_{p,\text{opt}}$: $\bm{M}_\text{opt} = \bm{M}_{p,\text{opt}} \bm{L}$.
\end{enumerate}

\begin{figure}[H]
\centering
\includegraphics[width=\textwidth]{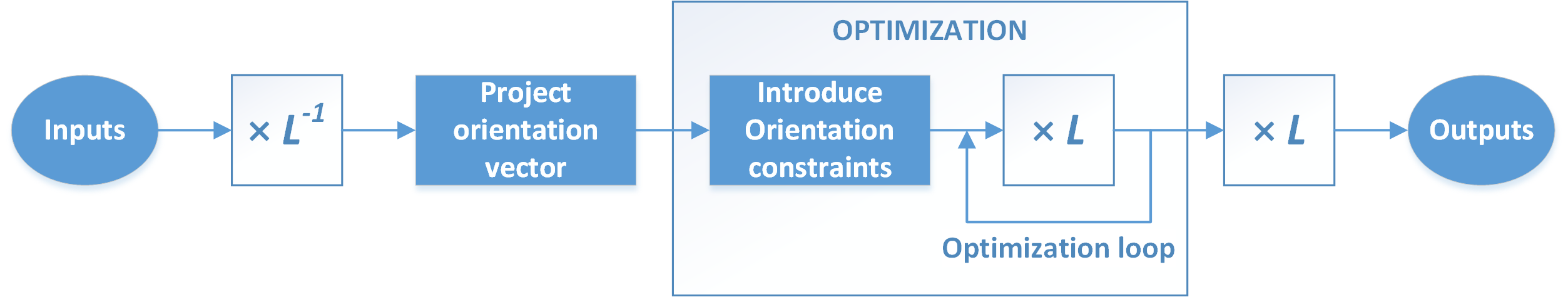}
\caption{Sequence of steps required to adapt the optimization problem with the orientation alignment. This adaptation is required in both the initial pose estimation and the final pose refinement.}
\label{fig:steps}
\end{figure}

To solve the optimization problem, we employed the same iterative method used in our previous work~\cite{Troncoso1,Troncoso2} based on Levenberg-Marquardt optimization~\cite{Levenberg,Marquardt}.
Figure~\ref{fig:transformations} shows a diagram of the different transformations that are performed during the detection process, including the original model and aligned model transformations.

\begin{figure}[H]
\centering
\includegraphics[width=0.75\textwidth]{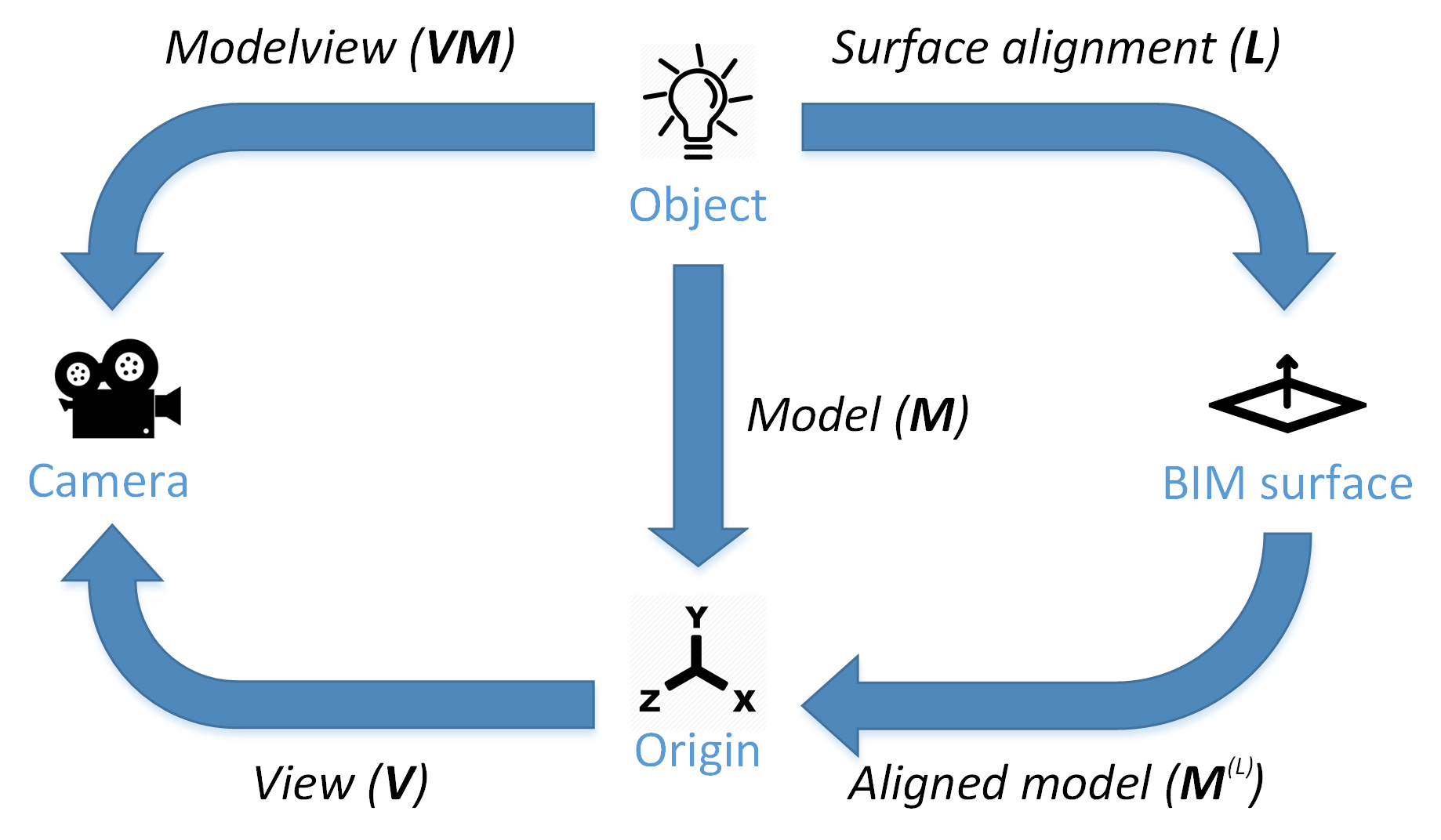}
\caption{Transformations between coordinate systems involved in the detection process.}
\label{fig:transformations}
\end{figure}

\subsection{Pose Estimation}

To force the orientation of the final detected poses, the first step was to modify the initial estimations based on the shapes in the image. We had to use different algorithms for polygonal and for circular shapes, as described below.

\subsubsection{Polygonal Shapes}

For polygonal shapes we solved the original perspective-n-point (PnP) problem as explained in~\cite{Troncoso1} to obtain the initial pose. Then, as described in Section~\ref{sec:const}, we transformed the pose and project the orientation vector to the plane normal. This first estimation was a good starting point for the second PnP problem, that tried to minimize the reprojection error of the object pose in the camera plane, using the traditional expression presented in (\ref{eq:spnp}), with $\bm{p}_i^{(o)}$ and $\bm{p}_i^{(c)}$ being the $i$-th pair of object and camera points, respectively:
\vspace{12pt}
\begin{align}\label{eq:spnp}
\min_{\bm{M}} &\quad \sum_{i} \left\|\bm{P}\bm{V}\bm{M}\bm{p}_i^{(o)} - \bm{p}_i^{(c)}\right\|^2 .
\end{align}

Using this optimization problem with the method in Section~\ref{sec:const} we obtain a pose that is aligned with the plane and has an expected low reprojection error.

\subsubsection{Circular Shapes}

For circular shapes, we used a modified version of the algorithm presented in~\cite{Troncoso2}, following analogous reasoning to that of Section~\ref{sec:const}. First, we changed the reference coordinate system with respect to~\cite{Troncoso2}: instead of the camera coordinates, we used the coordinates of the rotated object to be able to force the orientation of the normal vector.
Then, we had to project all the points in the object coordinate system to the coordinate system aligned with the plane normal. To denote this transformation, we use the following notation: $\bm{p}^{(L)} = \bm{L} \bm{p}$.

Using the equations of the projection line $\mathcal{L}$ from the camera origin to the image point, and the circle plane $\mathcal{P}$, we can obtain the projected point $\bm{p}_i'^{(L)}$ on $\mathcal{P}$ as in Equation (\ref{eq:cpnp1}), being $\bm{p}_c^{(L)}$ the camera center, $\bm{p}_C^{(L)}$ the circle center, $\bm{\hat n}$ the unit normal vector of $\mathcal{P}$, and $\bm{f}^{(L)} = \bm{p}_i^{(L)} - \bm{p}_c^{(L)}$:
\begin{align}\label{eq:cpnp1}
\begin{cases}
\mathcal{L}: \bm{p}_i'^{(L)} = \bm{p}_c^{(L)} + t \bm{f}^{(L)} \\
\mathcal{P}: \bm{\hat n}^{(L)} (\bm{p}_i'^{(L)} - \bm{p}_C^{(L)}) = 0
\end{cases} \rightarrow \quad t = \frac{\bm{\hat n}^{(L)} (\bm{p}_C^{(L)} - \bm{p}_c^{(L)})}{\bm{\hat n}^{(L)} \bm{f}^{(L)}} .
\end{align}

With Equation (\ref{eq:cpnp1}), and following the same procedure as the one presented in~\cite{Troncoso2}, we solved the minimization problem based on the distance to the circumference of radius $R_C$ to obtain the optimal values of $\bm{p}_C$ and $\bm{\hat n}$. This is presented in Equations (\ref{eq:cpnp2}) and (\ref{eq:cpnp2r}):
\begin{align}\label{eq:cpnp2}
\min_{\bm{p}_C^{(L)}, \bm{n}^{(L)}} &\quad \sum_{i} \left(\left\| \bm{p}_c^{(L)} + \frac{\bm{\hat n}^{(L)} (\bm{p}_C^{(L)} - \bm{p}_c^{(L)})}{\bm{\hat n}^{(L)} \bm{f}^{(L)}}\bm{f}^{(L)} - \bm{p}_C^{(L)} \right\| - R_C\right)^2 \\
\label{eq:cpnp2r}
\text{s.t.} &\quad \|\bm{\hat n}^{(L)}\| = 1 .
\end{align}

Finally, the resulting values of $\bm{p}_{C,\text{opt}}^{(L)}$ and $\bm{\hat n}_{\text{opt}}^{(L)}$ had to be projected back to the object coordinate system, as defined in Equations (\ref{eq:pnpl1}) and (\ref{eq:pnpl2}):
\begin{align}
\label{eq:pnpl1}
\bm{p}_{C,\text{opt}} =& \bm{L}^{-1}\bm{p}_{C,\text{opt}}^{(L)}\ ,\\
\label{eq:pnpl2}
\bm{\hat n}_{\text{opt}} =& \bm{L}^{-1}\bm{\hat n}_{\text{opt}}^{(L)}\ .
\end{align}

\subsection{Pose Refinement}

The refinement step must also be restricted to produce poses aligned with the appropriate surface in the BIM. In this case, the constraints are imposed on the input orientation vector for the direct directional chamfer optimization (D$^2$CO) method~\cite{D2CO}. In this case there is no difference between polygonal and circular shapes since the cost function is based on edge information from the image instead of point-to-point correspondences.

\subsection{Pose Filtering}

Forcing the orientation of the detected objects has the negative side effect of introducing potentially very different poses that do not match the original estimation. Therefore, an additional filter was required to discard these erroneous poses. We used the reprojection error to verify this similarity: the error function is given in Equation (\ref{eq:rep_err}) as the average squared reprojection error between the $N$ pairs of object and camera points $\{\bm{p}_i^{(o)}, \bm{p}_i^{(c)}\}$. The~shape area $A$ is introduced to normalize the error depending on the size and proximity of the object.
\vspace{12pt}
\begin{align}\label{eq:rep_err}
\varepsilon = \frac{1}{A} \frac{\sum_{i=1}^N \left\|\bm{P}\bm{V}\bm{M}\bm{p}_i^{(o)} - \bm{p}_i^{(c)}\right\|^2}{N} .
\end{align}

We can directly use (\ref{eq:rep_err}) with polygonal shapes; however, with circular shapes there are no direct correspondences of object points for the given camera points. Thus, for each camera point $\bm{p}_i^{(c)} = \left(x_i^{(c)}, y_i^{(c)}, z_i^{(c)}\right)$, we calculate the virtual object point $\bm{p}_i^{(o)} = \left(x_i^{(o)}, y_i^{(o)}, z_i^{(o)}\right)$ by choosing the closest point in the circumference to the projected camera point. We used the object coordinate system to simplify the expressions, performing the relevant transformations when necessary. The~sequence of operations is:

\begin{enumerate}
\item First, the projected camera point on the plane $z = 0$ was obtained, again, using the equations of the projection line $\mathcal{L}$ and the circle plane $\mathcal{P}$. This equation system is presented in Equation (\ref{eq:re1}), with $\bm{p}_c$ the camera center and $\bm{f} = \bm{p}_i^{(c)} - \bm{p}_c$:
\begin{align}\label{eq:re1}
\begin{cases}
\mathcal{L}: \bm{p}_i'^{(c)} = \bm{p}_c + t \bm{f} \\
\mathcal{P}: z_i'^{(c)} = 0
\end{cases} \rightarrow \quad t = -\frac{z_c}{z_f} .
\end{align}

\item The~intersection between the line and the circumference with radius $R_C$ was obtained by solving the system of equations in (\ref{eq:re2}), comprising the line $\mathcal{L}'$ from the circle center to $\bm{p}_i'^{(c)}$, and the circumference $\mathcal{C}$ of the object:
\begin{align}\label{eq:re2}
\begin{cases}
\mathcal{L}': x_i^{(o)} / x_i'^{(c)} = y_i^{(o)} / y_i'^{(c)} \\
\mathcal{C}: \left(x_i^{(o)}\right)^2 + \left(y_i^{(o)}\right)^2 = R_C^2
\end{cases} .
\end{align}

\item Choosing the result in the same quadrant gives the closest intersection that is used as the corresponding object point $\bm{p}_i^{(o)} = \left(x_i^{(o)}, y_i^{(o)}, 0\right)$ for $\bm{p}_i^{(c)}$.

\end{enumerate}

In our experiments, we used a threshold of 0.015 and 0.035 for polygonal and for circular shapes, respectively, based on experimental results.

\section{Experimental System}
\label{sec:exp}

The main contributions presented in Section~\ref{sec:method} were evaluated in five case studies, each with a different lamp model. These~five areas, shown in Figure~\ref{fig:areas}, correspond to the five lamp models of the database presented in~\cite{Troncoso2} that is used for the experiments. The~acquisition of the images was performed at a walking speed of $\approx$1 m/s, with a pitch of $\approx$60$^\circ$ with respect to the horizontal plane and positioning the camera at 1.5 m from the floor. The~1920 $\times$ 1080 images were extracted using a Lenovo Phab 2 Pro, with the Google Tango technology~\cite{Tango}. This acquisition protocol was the same as the one described in~\cite{Troncoso2}.

The first three case studies, corresponding to models 1 to 3 of the database, were located in the School of Industrial Engineering of the University of Vigo. The~reference values for these areas were obtained using manual inspection of the lamp poses. Case studies 4 and 5 were the same as the ones presented in~\cite{Troncoso2}, with data collected in the Mining and Energy Engineering School of the University of Vigo and ground truth values obtained from point clouds extracted with LiDAR sensors~\cite{Troncoso2}. The~number of images and lamps of the complete dataset for the experimental system are included in Table~\ref{tab:inventory}, with a total of more than 30,000 images.

The algorithms described in Section~\ref{sec:method} were implemented inside the same C++ framework developed for~\cite{Troncoso1, Troncoso2}, built using several software libraries to solve different problems related to image processing, geometry and optimization~\cite{OpenCV,OpenMesh,ceres,OpenGL}.

\begin{figure}[H]
\centering
\begin{subfigure}{0.325\linewidth}
\centering
\includegraphics[width=\textwidth]{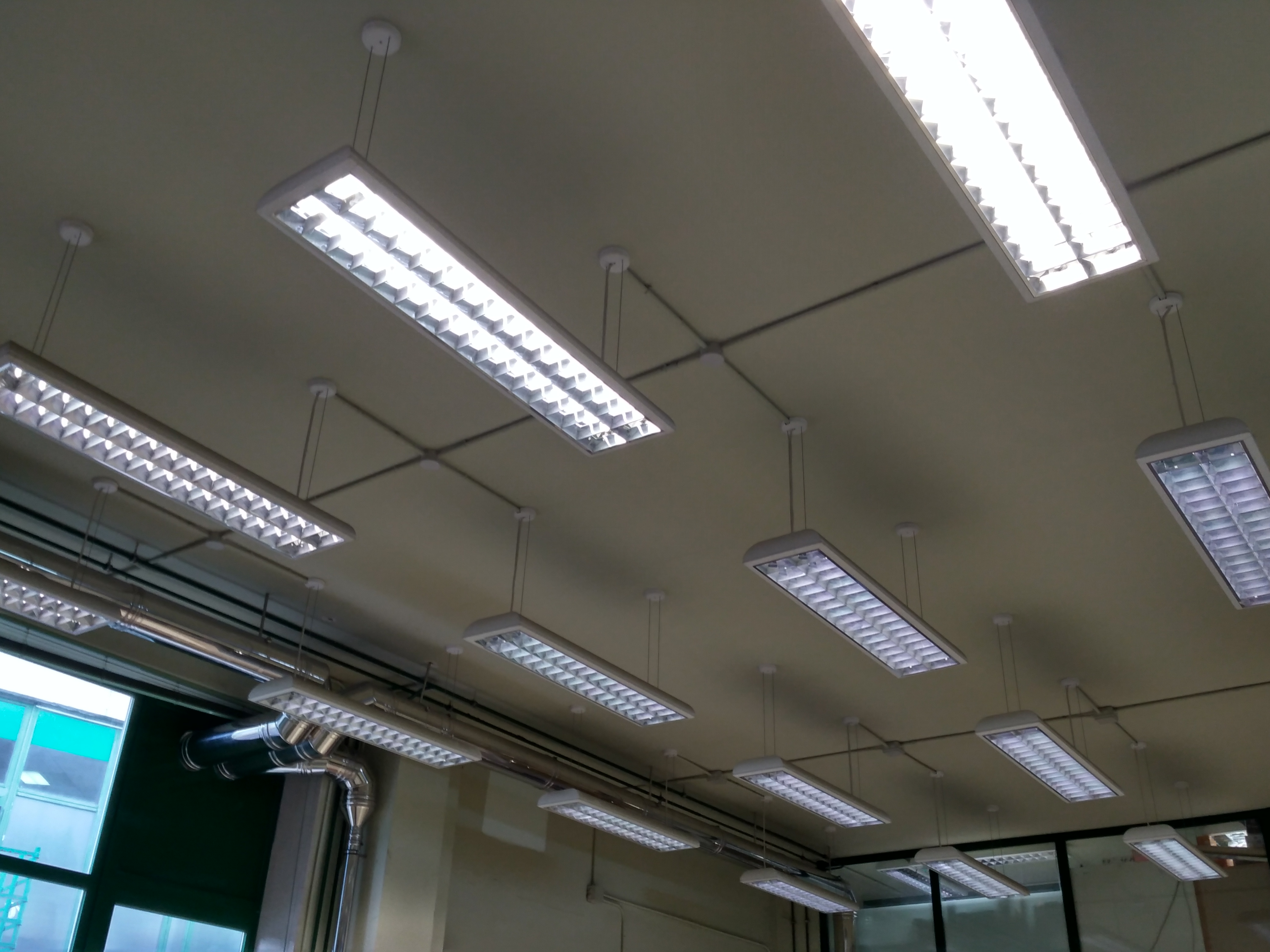}
\caption{}
\end{subfigure}
\begin{subfigure}{0.325\linewidth}
\centering
\includegraphics[width=\textwidth]{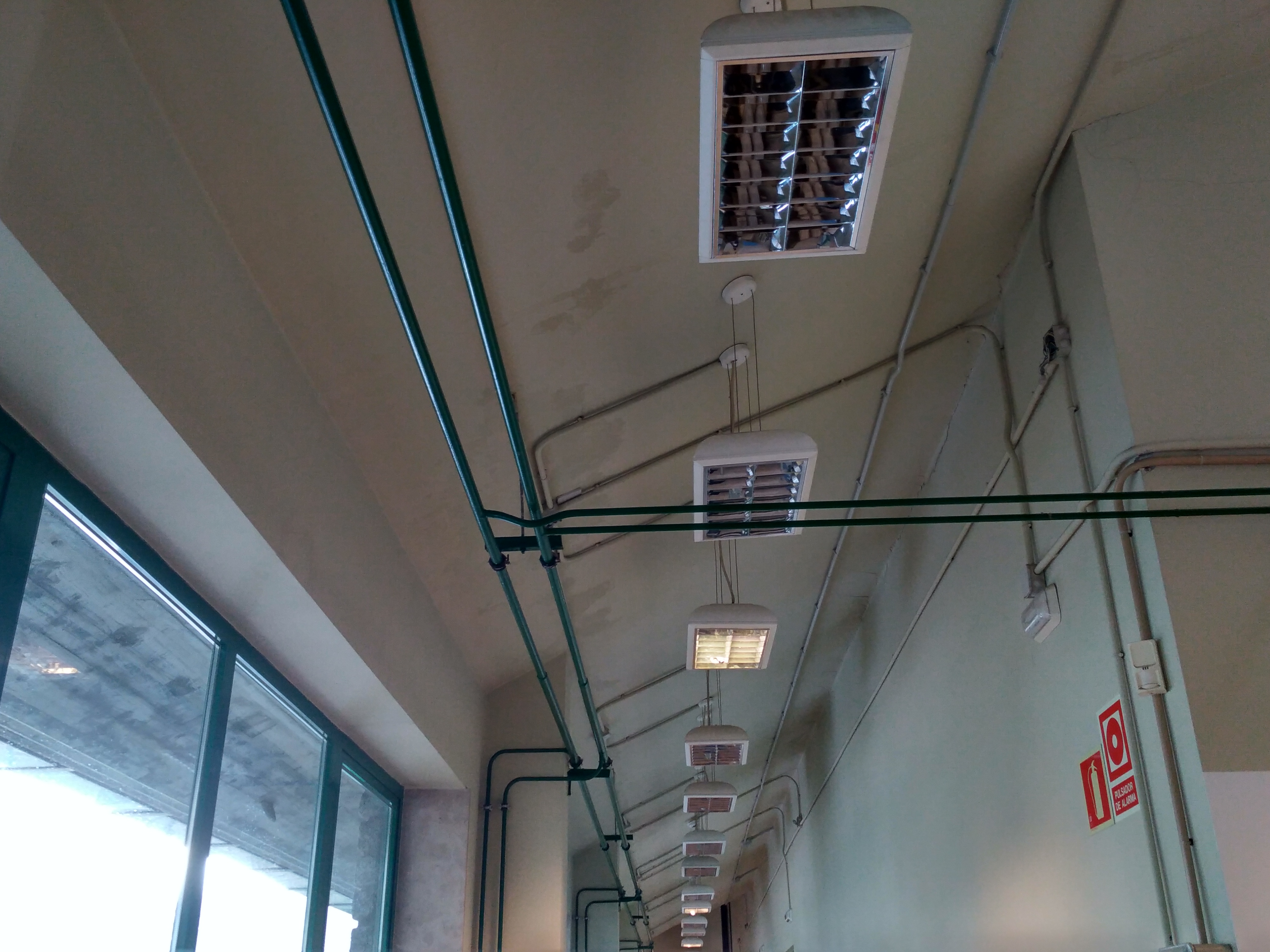}
\caption{}
\end{subfigure}
\begin{subfigure}{0.325\linewidth}
\centering
\includegraphics[width=\textwidth]{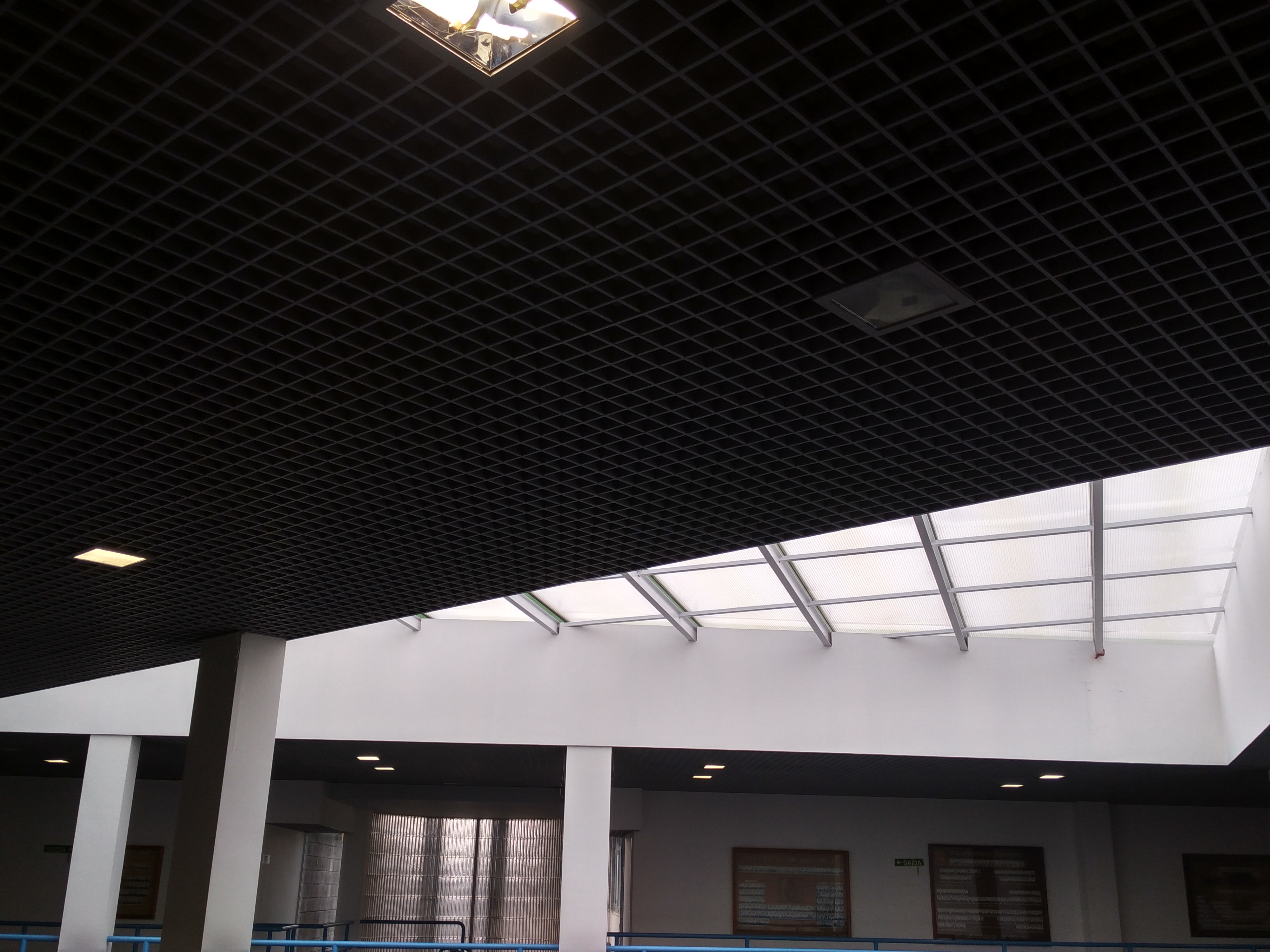}
\caption{}\vspace{6pt} 
\end{subfigure}
\begin{subfigure}{0.325\linewidth}
\centering
\includegraphics[width=\textwidth]{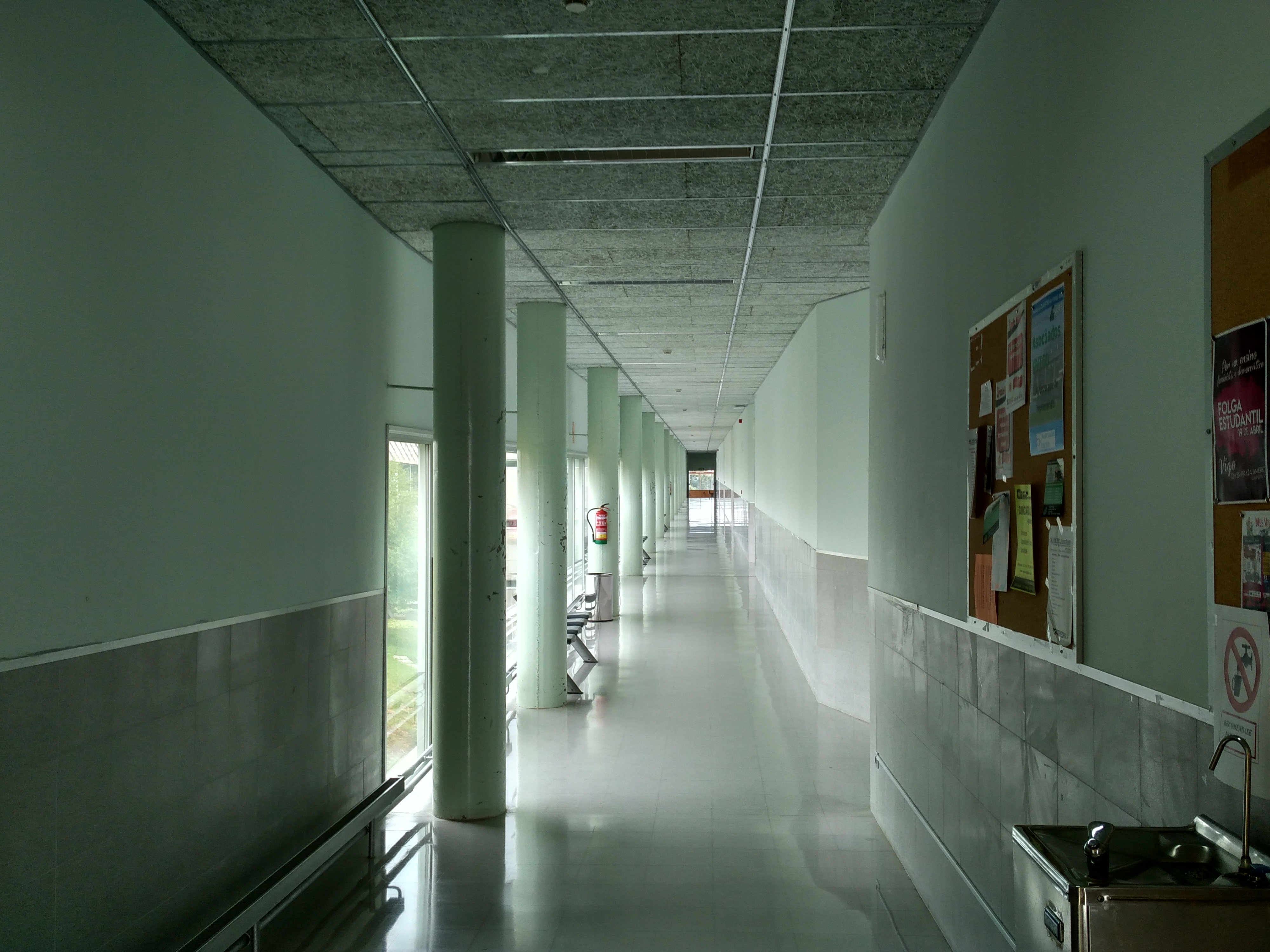}
\caption{}
\end{subfigure}
\begin{subfigure}{0.325\linewidth}
\centering
\includegraphics[width=\textwidth]{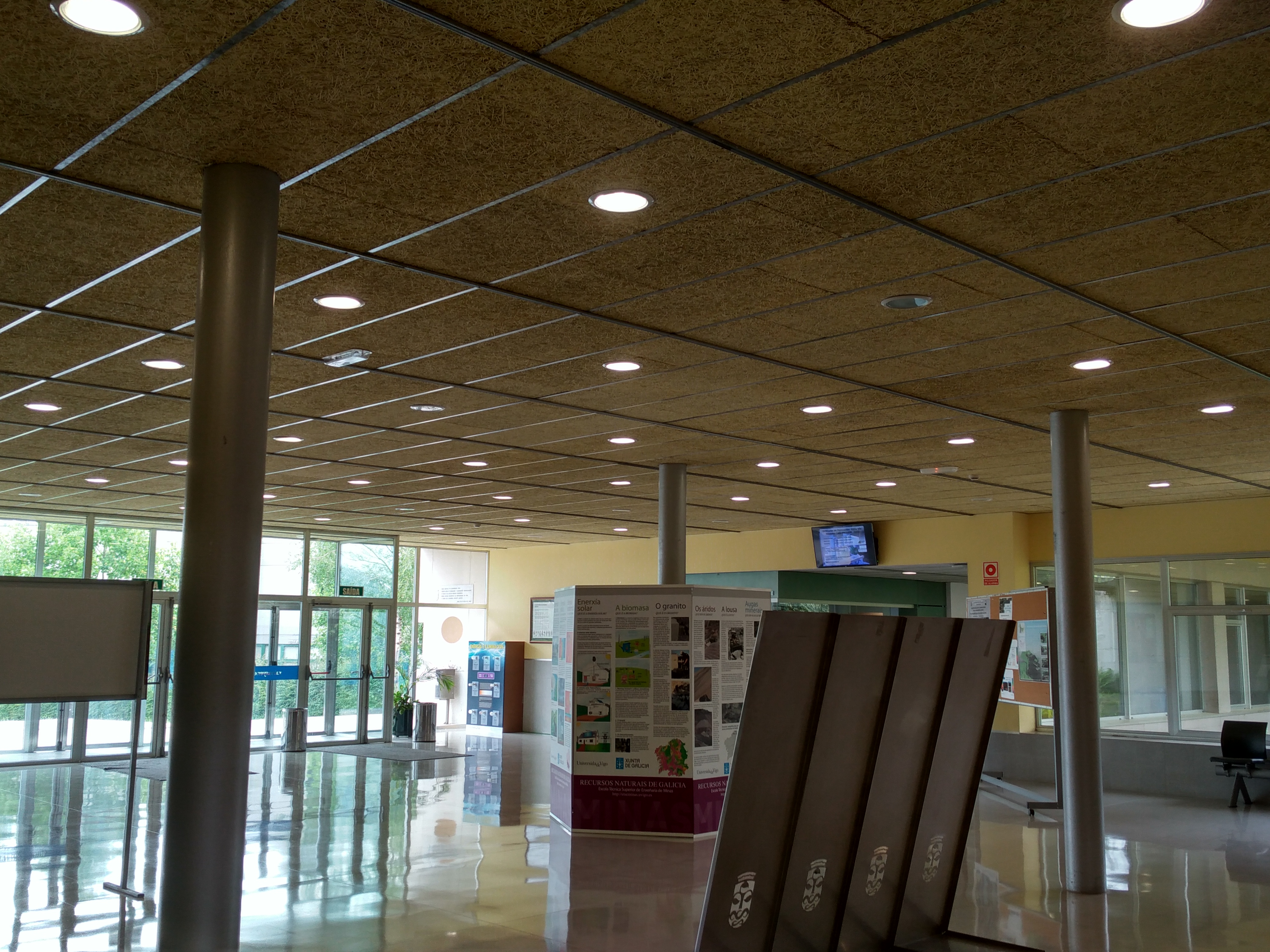}
\caption{}
\end{subfigure}
\caption{Spaces for the five case studies used to evaluate the system: ({\bf a}--{\bf e}) Case studies 1 to 5.} 
\label{fig:areas}
\end{figure}
\unskip

\begin{table}[H]
\centering
\small
\caption{Description of the complete dataset used, including a description of the physical space, the model number and the number of images, lamps and lamps turned on.}
\label{tab:inventory}
\begin{tabular}{p{8.3cm}cccccc}
\toprule
\rotatebox[origin=l]{0}{\textbf{Space Description}} &
\rotatebox[origin=l]{0}{\textbf{Model}} &
\rotatebox[origin=l]{0}{\textbf{No. Images}} &
\rotatebox[origin=l]{0}{\textbf{No. Lamps}} &
\rotatebox[origin=l]{0}{\textbf{No. on}} \\
\midrule
Laboratory, lamps suspended 50 cm from the ceiling, only two external windows, 1 m from the closest lamps & 1 & 5674 & 16 & 16 \\\midrule
Hallway, lamps suspended 40 cm from the ceiling, external windows at one side & 2 & 2453 & 19 & 10 \\\midrule
Reception, large open area, second floor, lamps fixed at the ceiling, bright environment & 3 & 2539 & 16 & 13 \\\midrule
Hallway, rectangular lamps embedded in the ceiling, external windows at one side & 4 & 6082 & 25 & 17 \\\midrule
Reception, circular lamps embedded in the ceiling & 5 & 14,535 & 90 & 67 \\
\midrule
\multicolumn{2}{l}{TOTAL} & 31,283 & 166 & 123 \\
\bottomrule
\end{tabular}
\end{table}

\section{Results and Discussion}
\label{sec:results}

\newcommand{\typestextA}{for the unconstrained system~\cite{Troncoso2}}
\newcommand{\typestextB}{with reprojection error filtering}
\newcommand{\typestextC}{with filtering and orientation alignment}
\newcommand{\typestext}{(\textbf{a}) Unconstrained system~\cite{Troncoso2}, (\textbf{b})~system with reprojection error filtering, and (\textbf{c}) system with filtering and orientation alignment}
\newcommand{\typestitle}{for the unconstrained system~\cite{Troncoso2}, the system with reprojection error filtering (REF), and the system with reprojection error filtering and orientation alignment (REF~+~OA)}
\newcommand{\typesheader}[1]{\multicolumn{#1}{c}{\textbf {Unconstrained}~\cite{Troncoso2}} & \multicolumn{#1}{c}{\textbf{REF}} & \multicolumn{#1}{c}{\textbf{REF~+~OA}}}

Using the experimental system described in Section~\ref{sec:exp}, we performed tests with three different versions of the system: (i) the unconstrained original system, that is the same as the one presented in~\cite{Troncoso2}, (ii) the system with the additional reprojection error filtering, and (iii) the complete improved system with reprojection error filtering and orientation alignment. Because of the introduction of the three new models, especially model 3 which had the lowest circularity~\cite{Rosin} values, we have modified the shape threshold for the experiments with respect to~\cite{Troncoso2}, using a value of 14, being this a good tradeoff between circular and polygonal shapes.

Example images with detections for each case study are presented in Figure~\ref{fig:exdets}. Moreover, the localized detections in the xy-plane are shown in Figures~\ref{fig:clus_1}--\ref{fig:clus_5} with the corresponding reference values. These~figures depict the final centers after the clustering operation, showing similar localization values with the three different versions of the system, with the most obvious improvement being the additional two correct detections for case study 5. These~results will be discussed in detail in the rest of the section, using three key metrics to evaluate the improvements; total number of detections, identification rate and distance to reference value.

\begin{figure}[H]
\centering
\begin{subfigure}{0.325\linewidth}
\centering
\includegraphics[width=\textwidth]{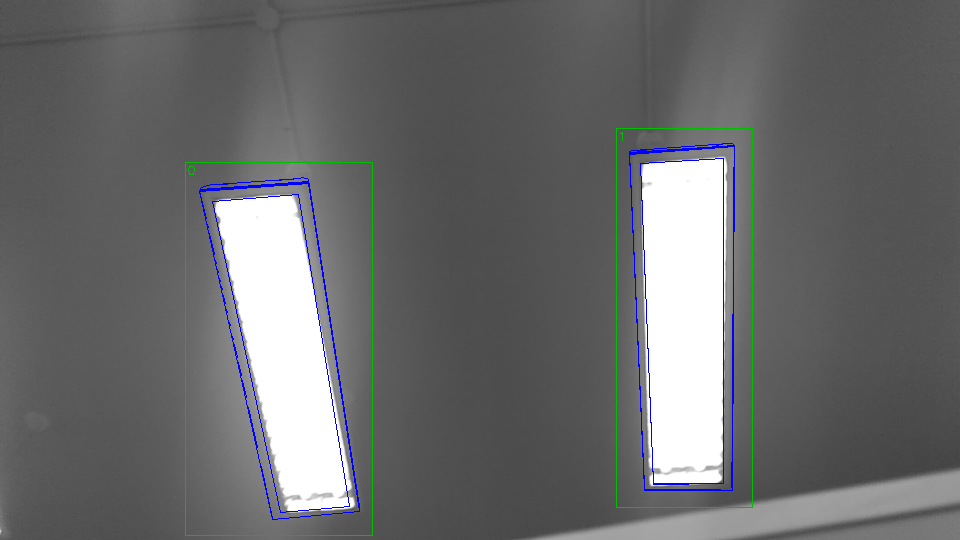}
\caption{}\vspace{6pt}
\end{subfigure}
\begin{subfigure}{0.325\linewidth}
\centering
\includegraphics[width=\textwidth]{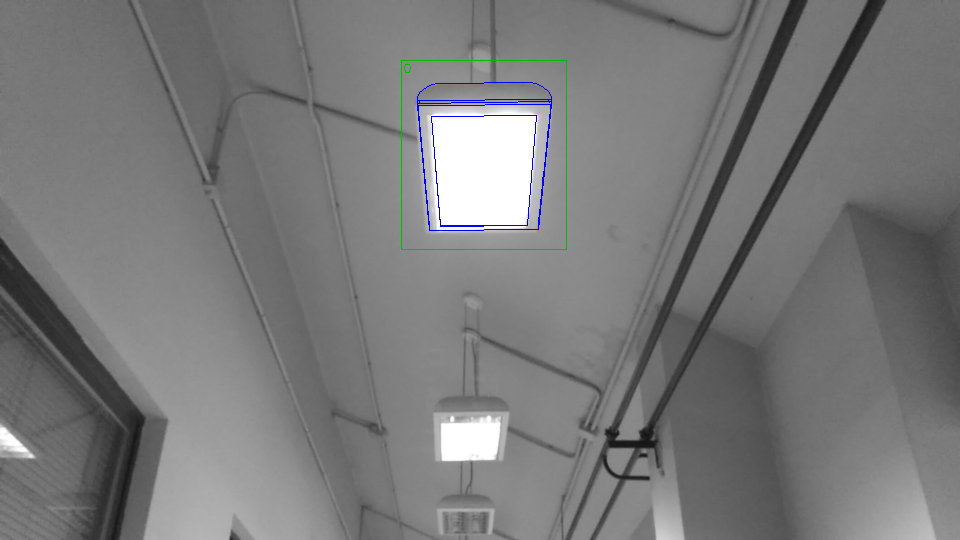}
\caption{}
\end{subfigure}
\begin{subfigure}{0.325\linewidth}
\centering
\includegraphics[width=\textwidth]{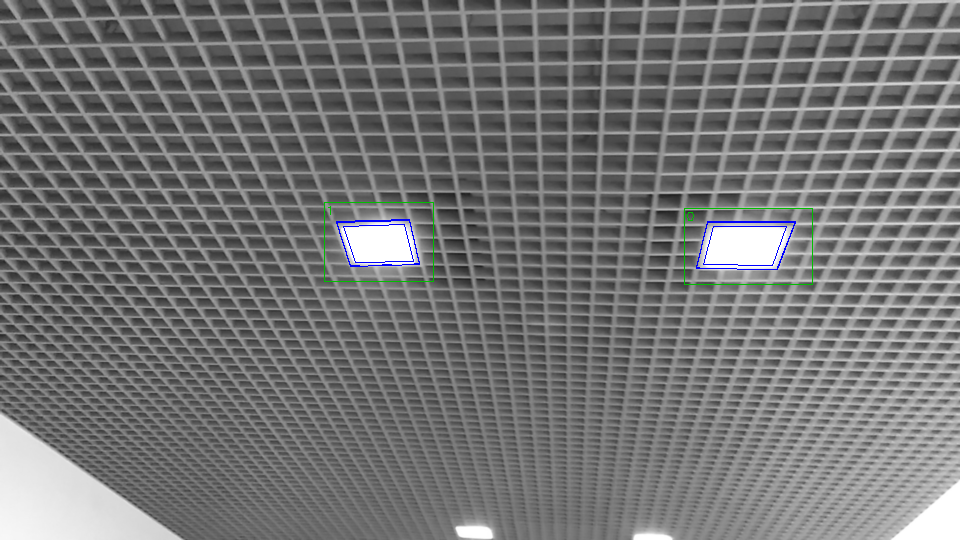}
\caption{}\vspace{6pt} 
\end{subfigure}
\begin{subfigure}{0.325\linewidth}
\centering
\includegraphics[width=\textwidth]{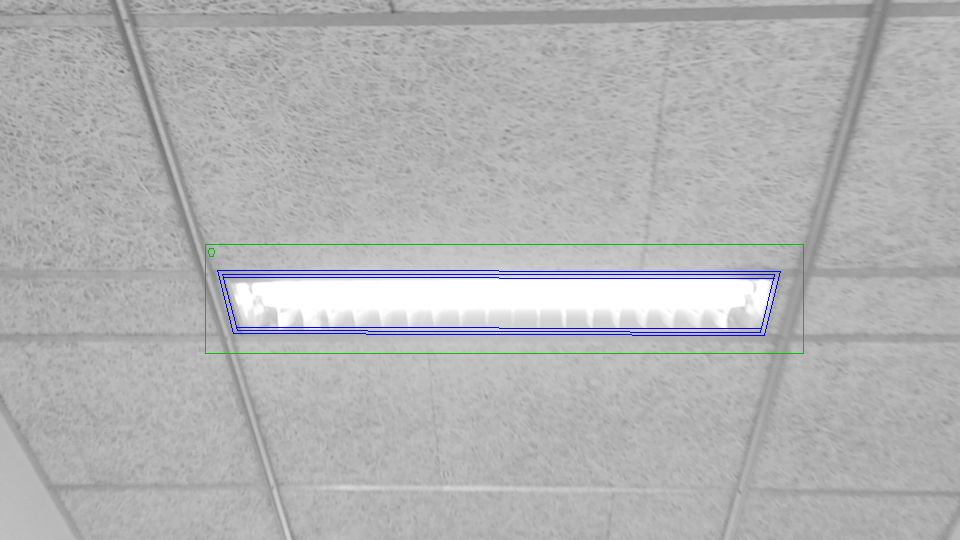}
\caption{}
\end{subfigure}
\begin{subfigure}{0.325\linewidth}
\centering
\includegraphics[width=\textwidth]{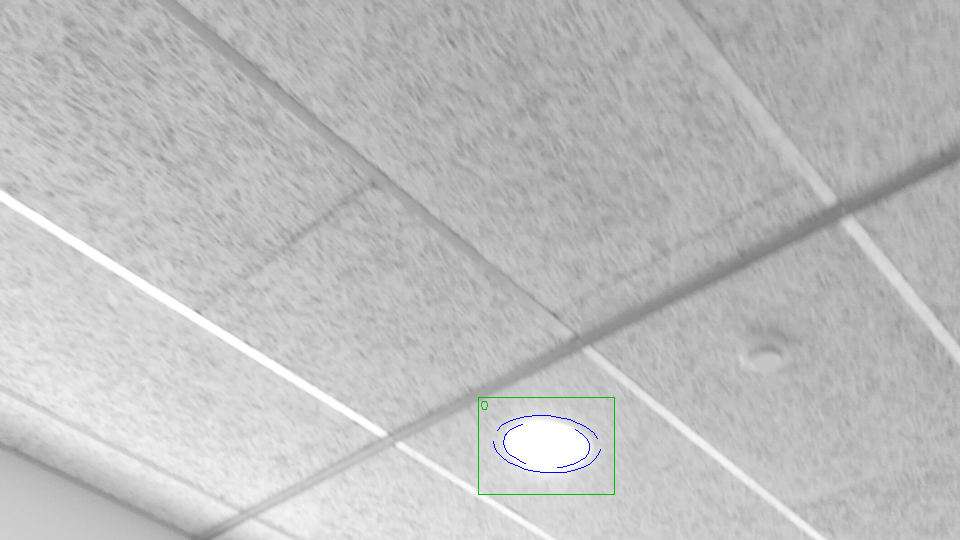}
\caption{}
\end{subfigure}
\caption{Detections for each of the case studies in the experiments: ({\bf a}--{\bf e}) Case studies 1 to 5.} 
\label{fig:exdets}
\end{figure}

\newcommand{\clusimgs}[1]{
\begin{figure}[H]
\centering
\begin{subfigure}{0.325\linewidth}
\centering
\includegraphics[width=\textwidth]{img/clus_#1_0}
\caption{}
\end{subfigure}
\begin{subfigure}{0.325\linewidth}
\centering
\includegraphics[width=\textwidth]{img/clus_#1_1}
\caption{}
\end{subfigure}
\begin{subfigure}{0.325\linewidth}
\centering
\includegraphics[width=\textwidth]{img/clus_#1_2}
\caption{}
\end{subfigure}
\caption{Cluster centers and reference values for case study #1. \typestext.}
\label{fig:clus_#1}
\end{figure}
}
\vspace{-6pt}

\clusimgs{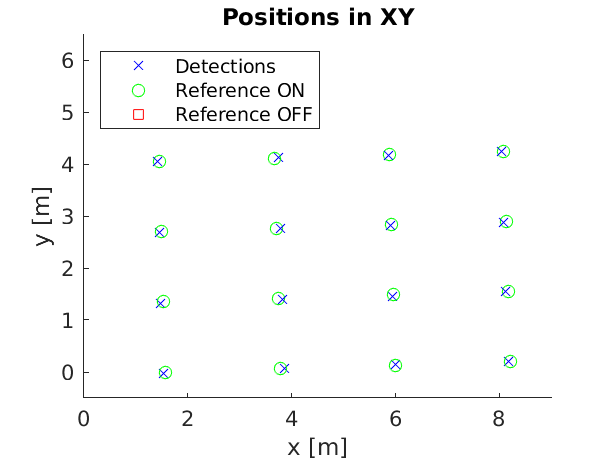}
\vspace{-6pt}
\clusimgs{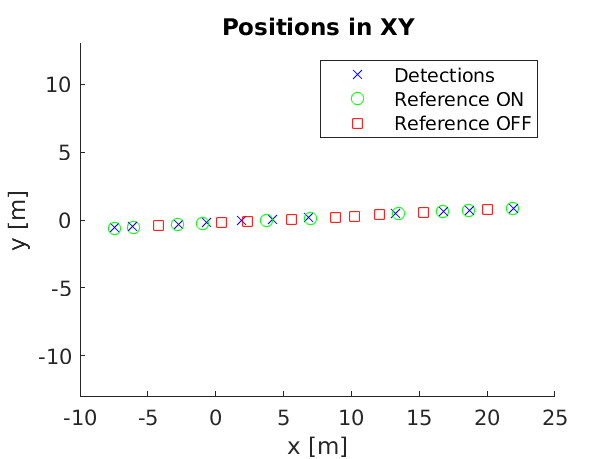}
\vspace{-6pt}
\clusimgs{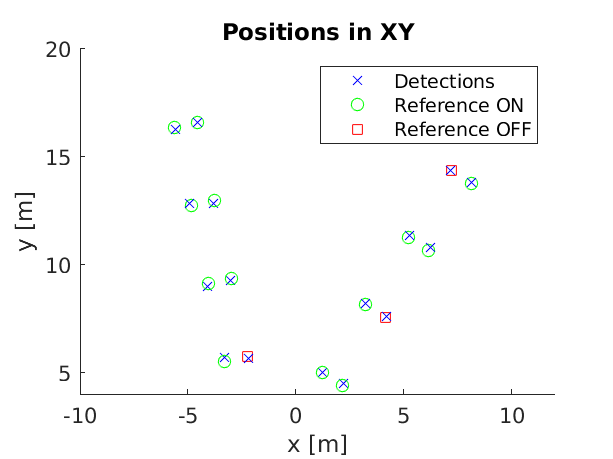}
\vspace{-6pt}
\clusimgs{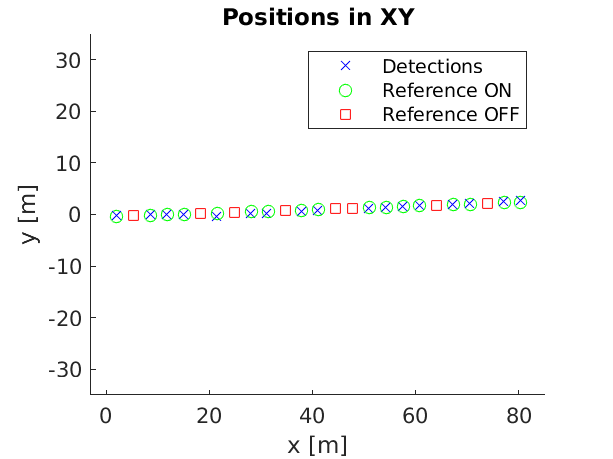}
\vspace{-6pt}
\clusimgs{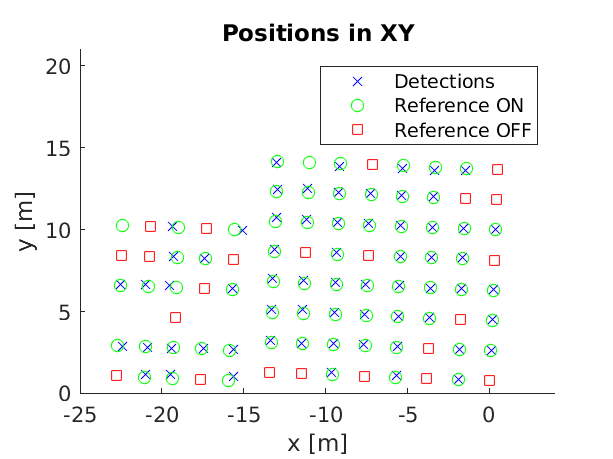}

\subsection{Number of Detections}

The total number of detections for the three different modes are presented in Table~\ref{tab:counts}. The~use of the reprojection error filtering lowered the number of detections even when the orientation alignment is not used, removing detections whose projected shape on the image is too different from the expected values. The~orientation alignment, however, greatly increases the total number of detections, even with the filtering, with an increase of 48.91\%. This resulted in a more reliable detection system, as the probability of capturing the object is higher, specially when it is only visible in a small number of~frames.

The results per cluster evidence the importance of increased reliability, as shown in Table~\ref{tab:cluster-stats}. This~table shows relevant figures for the number of detections that are used to calculate each cluster center. We can see that there is a great increase in the minimum number of detections, with the original system having only one detection for one cluster in case study 4, while the updated system always has at least 18 detections per cluster.

\begin{table}[H]
\centering
\small
\caption{Total number of detections \typestitle.}
\begin{tabular}{cccc}
\toprule
\textbf{Case Study} & \typesheader{1}\\
\midrule
1&1810&1811&2421\\
2&  702&  701&  840\\
3&  735&  707&  701\\
4&  670&  666&2426\\
5&4743&4733&6421\\
\midrule
TOTAL&8660&8618&12,809\\
&100\%&99.52\%&148.91\%\\
\bottomrule
\end{tabular}
\label{tab:counts}
\end{table}
\vspace{-6pt}

\begin{table}[H]
\centering
\small
\caption{Statistics of the number of detections per cluster \typestitle.}
\begin{tabular}{cccc ccc ccc}
\toprule
\textbf{Case Study} & \typesheader{3}\\
& \textbf{Min} & \textbf{Mean} & \textbf{Max}
& \textbf{Min} & \textbf{Mean} & \textbf{Max}
& \textbf{Min} & \textbf{Mean} & \textbf{Max}\\
\midrule
1&17&113.13&303&17&113.19&303&35&151.31&336\\
2&11& 63.82&146&11& 63.73&146&18& 76.36&157\\
3&20& 45.94& 80&20& 44.19& 80&19& 43.81& 87\\
4& 1& 39.41&174& 1& 39.18&174&74&142.71&187\\
5& 2& 72.97&169& 2& 72.82&169&31& 95.84&179\\
\midrule
GLOBAL
&1&67.05&174.4
&1&66.62&174.4
&18&102.01&189.2\\
&100\%&100\%&100\%
&100\%&99.35\%&100\%
&1800\%&152.13\%&110.89\%\\
\bottomrule
\end{tabular}
\label{tab:cluster-stats}
\end{table}

\subsection{Identification Rate}

The correct identification of lamp model and state is crucial for the system. Figures~\ref{fig:distrib_0}--\ref{fig:distrib_2} show the distribution of accumulated lamp model scores identified for each cluster. Each of the bars corresponds to the total sum of scores for the individual detections, with different colors for each lamp model and the final decision for the cluster below.

The additional filtering had a very small impact on the total scores, while the use of the orientation alignment greatly increased the values. As previously mentioned, this is important to guarantee good results in the final clusters, and is specially relevant in some clusters, particularly for case studies 4 and 5, that had low values with the unconstrained system~\cite{Troncoso2}.

Based on these accumulated scores, the system was able to identify the correct model for all the clusters independently of the method used, but the number of correct individual identifications is, again, important for the reliability of the identification performance of the system. To provide a more direct analysis of this aspect, the confusion matrices for lamp model and lamp state are presented in Figures~\ref{fig:conf_0} and~\ref{fig:conf_1}, respectively. Here, both the filtering and the alignment had positive effects in the model identification, keeping a very low error rate with orientation alignment even when there are far more detections. The~total error decreases from 0.30\% with the unconstrained system, to 0.17\% reprojection error filtering and, finally, to 0.07\% with the complete new system.

Regarding lamp state, the filtering had no effect on the final error value of 3.61\%, but the orientation alignment yields two additional clusters corresponding to lamps that are turned on, improving the final error value to 2.41\%.

\newcommand{\distribimgs}[2]{
\begin{figure}[H]
\centering
\begin{subfigure}{0.325\linewidth}
\centering
\includegraphics[width=\textwidth]{img/distrib_1_#1}
\caption{}
\end{subfigure}
\begin{subfigure}{0.325\linewidth}
\centering
\includegraphics[width=\textwidth]{img/distrib_2_#1}
\caption{}
\end{subfigure}
\begin{subfigure}{0.325\linewidth}
\centering
\includegraphics[width=\textwidth]{img/distrib_3_#1}
\caption{}\vspace{6pt} 
\end{subfigure}
\begin{subfigure}{0.325\linewidth}
\centering
\includegraphics[width=\textwidth]{img/distrib_4_#1}
\caption{}
\end{subfigure}
\begin{subfigure}{0.665\linewidth}
\centering
\includegraphics[width=\textwidth]{img/distrib_5_#1}
\caption{}
\end{subfigure}
\caption{Accumulated scores for each lamp model in each cluster of the five case studies #2. The~model corresponding to the highest value is included below each bar: ({\bf a}--{\bf e}) Case studies 1 to 5.}
\label{fig:distrib_#1}
\end{figure}
}
\vspace{-6pt}

\distribimgs{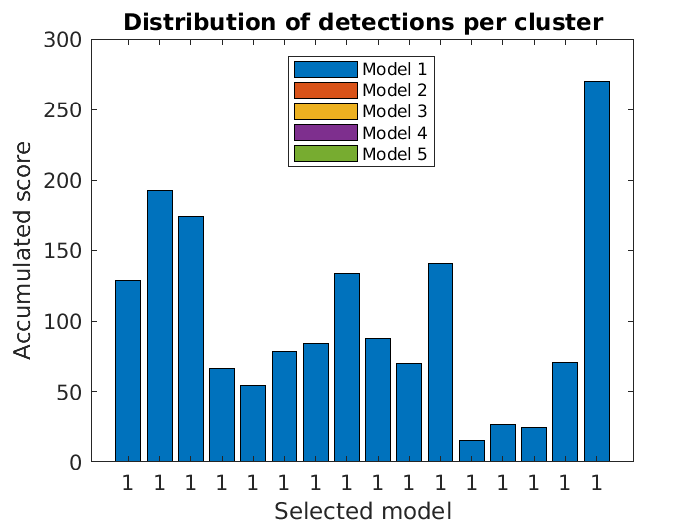}{\typestextA}
\vspace{-6pt}
\distribimgs{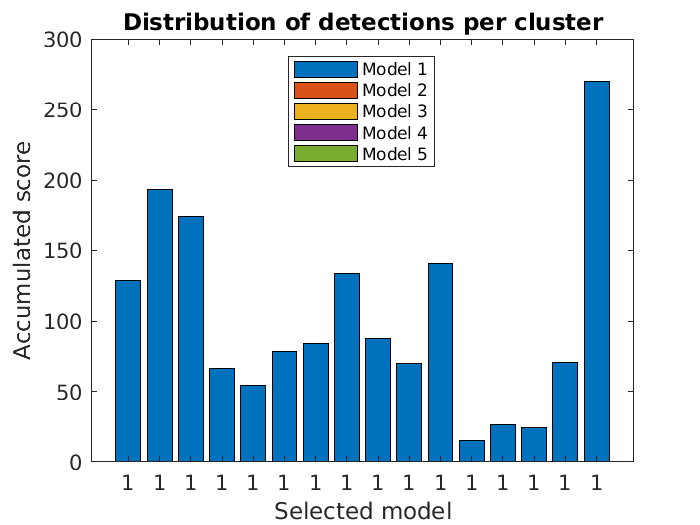}{\typestextB}
\vspace{-6pt}
\distribimgs{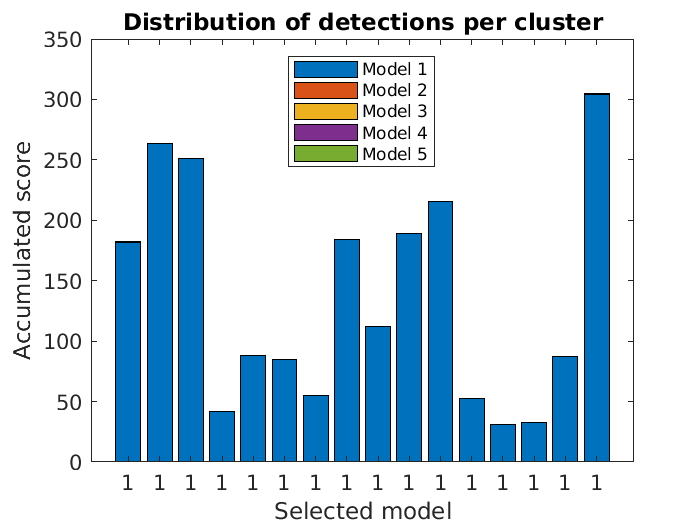}{\typestextC}
\vspace{-6pt}
\newcommand{\confimgs}[2]{
\begin{figure}[H]
\centering
\begin{subfigure}{0.327\textwidth}
\centering
\includegraphics[width=\textwidth]{img/conf_0_#1}
\caption{}
\end{subfigure}
\begin{subfigure}{0.327\textwidth}
\centering
\includegraphics[width=\textwidth]{img/conf_1_#1}
\caption{}
\end{subfigure}
\begin{subfigure}{0.327\textwidth}
\centering
\includegraphics[width=\textwidth]{img/conf_2_#1}
\caption{}
\end{subfigure}
\caption{Confusion matrices #2. Values in the diagonal indicate the number of correct matches for each class, while the rest of the values correspond to incorrect identifications depending on the expected and detected class. Percentages of correct and incorrect identifications are included in the last row and column. \typestext}
\label{fig:conf_#1}
\end{figure}
}

\confimgs{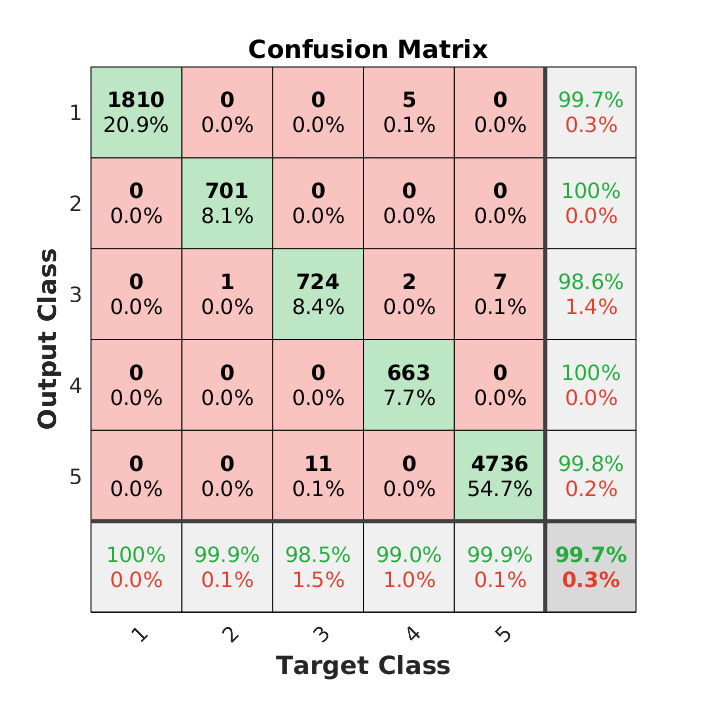}{for the model classes}
\vspace{-6pt}

\confimgs{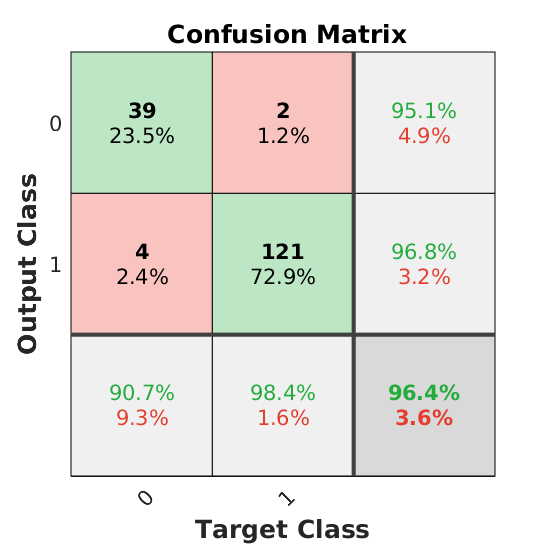}{for the lamp state, with class 0 and 1 representing the off and on state, respectively}

\subsection{Distance to Reference}

The final locations of the cluster centers have to match the reference positions to provide an accurate localization of the lamps. While Figures~\ref{fig:clus_1}--\ref{fig:clus_5} provide a visual representation of this similarity, Table~\ref{tab:rdis} contains numeric values for the average distances between the position of each cluster center and its corresponding closest reference point. The~inclusion of the additional filter provides a marginally worse localization performance, from 14.14 cm to 14.24 cm, but the use of the pose alignment improves the results, lowering the average distance to 13.63 cm.

\begin{table}[H]
\centering
\small
\caption{Average distance between cluster centers and reference positions, in centimeters, \typestitle.}
\begin{tabular}{cccc}
\toprule
\textbf{Case Study} & \typesheader{1}\\
\midrule
1& 4.7663& 4.7645& 4.7811\\
2&17.7931&17.8225&17.3197\\
3& 9.0811& 9.4964& 9.8789\\
4&25.9497&26.0241&25.0381\\
5&13.1264&13.1128&11.1107\\
\midrule
TOTAL&14.1433&14.2441&13.6257\\
&100\%&100.71\%&96.34\%\\
\bottomrule
\end{tabular}
\label{tab:rdis}
\end{table}

\subsection{Applications and Future Work}

The results presented in this section show that the proposed system is valid for the intended use cases, with an increased performance with respect to previous methods. The~required steps to leverage this new system include the collection of (i) the geometry of, at least, the ceiling of the spaces that contains the lamps and (ii) a database containing, but not limited to, the expected 3D lamp models, which can be obtained from manufacturers.
Moreover, special care must be taken with hanging lamps in the last projection step. In this work, we have used the distance from the ceiling to the lamp plane to generate the final results to validate the contributions, but this variable might not be known~beforehand.

Thus, to improve the system and provide a more streamlined solution, we are currently working on methods to estimate this virtual plane automatically from the detection points and the BIM information, removing the requirement of any prior knowledge regarding the configuration of lamp~positions.

\section{Conclusions}
\label{sec:conclusions}

In this work, we have presented two new contributions to our previous system for the detection of lighting elements in buildings: (i) the early use of BIM information to restrict the possible orientation values of the detections and (ii) a reprojection error filter that discards poses that do not match the estimated light surface shape. The~new constrained system was tested with a dataset of more than 30,000 images in five case studies with a total of 166 lamps of different models to analyze the quantitative improvements of the proposed modifications.

First, the number of individual detections is increased from 8618 to 12,809, almost 50\% higher, making the system more reliable, specially for clusters with a low detection count on the original system. Moreover, the number of incorrect model identifications is reduced from 0.30\% to 0.07\%, preserving a very low number of errors despite the high increase in the number of detections. Furthermore, the identification of lamp state is also improved, with the error decreasing from 3.61\% to 2.41\%. Finally, the average distance between the cluster centers and the reference positions is reduced from 14.14 cm to 13.63 cm. These~results shows an improvement of the new system with all the metrics used, yielding better detection rate, identification performance and localization accuracy.

\vspace{6pt}

\section{Acknowledgements}

Authors want to give thanks to the Xunta de Galicia under Grant ED481A and the Spanish Ministry of Economy and Competitiveness
under the National Science Program TEC2017-84197-C4-2-R.

\section{Abbreviations}
The following abbreviations are used in this manuscript:\\

\noindent
\begin{tabular}{@{}ll}
BIM & Building information modelling\\
IFC & Industry foundation classes\\
gbXML & Green Building XML Schema\\
LED & Light emitting diode\\
CVS & Computer vision system\\
SIFT & Scale-invariant feature transform\\
SURF & Speeded-up robust features\\
BOLD & Bunch of lines descriptor\\
BORDER & Bounding oriented-rectangle descriptors for enclosed regions\\
BIND & Binary integrated net descriptors\\
FDCM & Fast directional chamfer matching\\
D$^2$CO & Direct directional chamfer optimization\\
ROI & Region of interest\\
PnP & Perspective-n-point\\
\end{tabular}


\bibliographystyle{ieeetr}





\end{document}